\documentclass{article}
\usepackage{ijcai15}
\usepackage{times}

\usepackage{epsfig}
\usepackage{graphicx}
\usepackage{amsmath}
\usepackage{amssymb}
\usepackage{dsfont}
\usepackage{subfigure}
\usepackage{multirow}
\usepackage{url}
\usepackage{balance}
\usepackage{sidecap}
\usepackage{array}
\usepackage{natbib}
\usepackage{sidecap}

\usepackage{colortbl}

\usepackage{booktabs,xcolor,siunitx}
\definecolor{lightgray}{gray}{0.9}

\newcolumntype{V}{>{$\vcenter\bgroup\hbox\bgroup}c<{\egroup\egroup$}}
\def\Hline{\noalign{\hrule height 4\arrayrulewidth}}

\graphicspath{{./imgs/}}

\title{Highly Accurate Gaze Estimation Using a Consumer RGB-D Sensor}
\author{~\begin{tabular}{cc}
Reza Shoja Ghiass & Ognjen Arandjelovi\'c\\
Computer Vision and Systems Laboratory & School of Computer Science\\
Universit\'{e} Laval    & University of St Andrews\\
Qu\'{e}bec (QC) G1V 0A6 & St Andrews, KY16 9SX\\
Canada & United Kingdom\\
       & \small\texttt{ognjen.arandjelovic@gmail.com}\\
\end{tabular}~
}

\begin{document}

\maketitle

\begin{abstract}
Determining the direction in which a person is looking is an important problem in a wide range of HCI applications. In this paper we describe a highly accurate algorithm that performs gaze estimation using an affordable and widely available device such as Kinect. The method we propose starts by performing accurate head pose estimation achieved by fitting a person specific morphable model of the face to depth data. The ordinarily competing requirements of high accuracy and high speed are met concurrently by formulating the fitting objective function as a combination of terms which excel either in accurate or fast fitting, and then by adaptively adjusting their relative contributions throughout fitting. Following pose estimation, pose normalization is done by re-rendering the fitted model as a frontal face. Finally gaze estimates are obtained through regression from the appearance of the eyes in synthetic, normalized images. Using EYEDIAP, the standard public dataset for the evaluation of gaze estimation algorithms from RGB-D data, we demonstrate that our method greatly outperforms the state of the art.
\end{abstract}

\section{Introduction}\label{s:intro}
The need to know the direction of gaze of a person is a challenge encountered in many human centred computer applications. It is of pervasive interest in marketing \citep{HorsElioKnigReil2014}, in human-computer interaction \citep{YuanZhaoTuShao2011}, gaming \citep{CorcNanuPetrBigi2012}, psychological research \citep{BaOdob2009}, face recognition \citep{Aran2012b,GhiaAranBendMald2013a,AranHammCipo2010}, and many others. Therefore it is unsurprising that the problem of inferring gaze is a popular and well established research topic in computer vision which continues to challenge the state of the art \citep{HansJi2010}.

Most of the published methods on gaze estimation precede the emergence of cheap and readily available depth sensors such as those addressed in the present paper. Therefore these, which we shall for the sake of brevity refer to as `conventional' approaches, rely purely on visual (in general colour or more often simply pixel intensity) information. Amongst these conventional approaches two broad classes of approaches can be recognized: (i) model based, and (ii) learning based. The former group of methods uses an explicit 3D model of the eye to estimate gaze direction. Almost invariably methods of this group require calibration which is a significant practical limitation, in that it is cumbersome and tedious to the user. Recent and notable methods of this group include those of \cite{YamaUtsuYoneAbe2008,YangSunLiuLi+2012,Taba2012,SiguSidh2011,HungYin2010,ModeEize2010,NagaSugaIwamKama+2010}. For further detail the reader is directed to a comprehensive survey recently performed by \cite{HansJi2010}.

The second major group of conventional approaches adopts a more explicit, learning based model. Generally speaking algorithms of this type attempt to learn the mapping from the space of eye appearance images to the space of screen gaze points or gaze directions. Similar in their general approach, methods of this type exhibit differences in terms of eye appearance representation and the specific statistical models employed to learn the aforementioned mapping. Notable methods include those describe by \cite{TanKrieAhuj2002,SheeVija2011,OrozRocaGonz2009,LuSugaOkabSato2011a,SugaMatsSato2012,CoutMori2012}.

Notwithstanding this continued major research effort, practical gaze estimation remains a significant research challenge. In particular, the competing requirements of usability, accuracy, and robustness, amongst others, have proven difficult to achieve. Recent advances in the availability and affordability of sensors of alternative modalities in the consumer market offer if not a possible solution, then certainly a major source of potential improvement in the aforementioned aspects of gaze estimation systems. In particular in the present work we are interested in inferring gaze direction using a combination of conventional RGB image data and low quality, noisy depth data provided by devices such as Microsoft Kinect. This problem has so far received little attention, save for the work by \cite{FuneOdob2012} which is the current state of the art.

\section{Proposed method}
On the coarsest level the method we introduce in this paper comprises two stages. In the first stage the sensed depth data is used to reconstruct a person specific 3D model of the user's face, and then to create a synthetic frontal image by simulating a rigid 3D transformation of the face and its re-rendering. This normalizing step is used to constrain the subsequent learning stage which estimates the user's point of gaze by regression. The two stages of the algorithm are explained in detail next.

\begin{figure*}
  \centering
  \includegraphics[width=\textwidth]{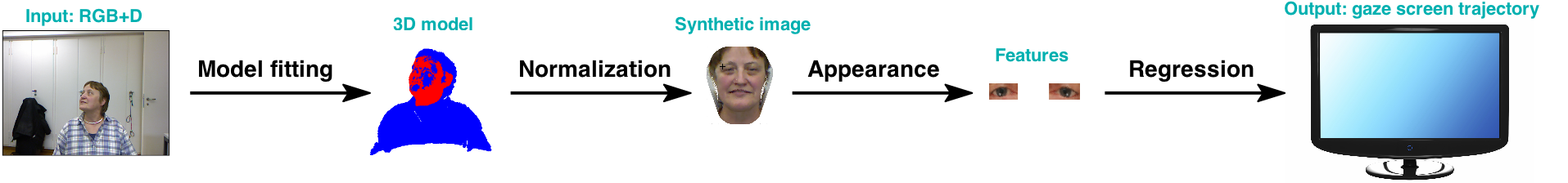}
  \caption{A schematic summary of the main steps of the proposed algorithm.}
  \label{f:overview}
\end{figure*}

\subsection{Pose estimation}
Unlike much of the previous work we perform pose estimation explicitly, that is, the three pose parameters (Euler angles) are explicit parameters of the underlying model rather than inferred through regression. This removes the need for elaborate training which necessitates extensive and laboriously labelled data. We use the so-called 3D morphable model~\citep{SunYin2008} which we fit directly to the sensed depth data as by \cite{GhiaAranLaur2015}, that is, without the use of RGB appearance (unlike e.g.\ \cite{BlanVett1999}). The fitting is formulated as an optimization task solved iteratively using a type of iterative closest point algorithm~\citep{WollAran2011,BouaTaglPaul2013}.

\subsubsection{Basic fitting framework}
In this first stage of the proposed method our goal is to fit a 3D morphable model~\citep{RomdBlanVett2002,RomdVett2003} to the sensed (``target'') point cloud $Y=\{y_1,\ldots,y_m\}$. The 3D morphable model captures shape variation through a linear combination of the principal shapes each of which is a dense triangulated mesh of vertices which correspond to identical anatomical and semantic loci across faces. A shape vector $\mathbf{s}$ which contains stacked 3D coordinates of model vertices can be written as:
\begin{align}
  \mathbf{s}(\theta) = \mu_s + \mathbf{S}\theta
\end{align}
where $\mu_s$ is the mean face shape, $\mathbf{S}$ a column matrix of shape basis vectors, and $\theta$ the set of parameters of the model i.e.\ the coefficients associated with the shape basis vectors~\cite{BlanVett1999}.

Herein the task of fitting is posed as a registration problem whereby the aim is to register the ``source''  point cloud $X=\{x_1,\ldots,x_n\}$, generated by sampling from the 3D morphable model synthesised surface, with the target by minimizing an error function which comprises a weighted summation of three terms:
\begin{align}
  E_\text{fit} = E_\text{match} + \omega_1~E_\text{rigid}+\omega_2~E_\text{model}.
  \label{e:objFn}
\end{align}
The first term, $E_\text{match}$, quantifies the proximity of source and target point clouds. The second term, $E_\text{rigid}$, seeks to impose rigidity of registration by penalizing point correspondences between the two clouds which correspond to non-rigid deformations. Lastly $E_\text{model}$ penalizes unlikely intrinsic model parameters. The parameters $\omega_1$ and $\omega_2$ determine the relative contributions of the three error terms.

To increase its robustness to incorrect or misleading point cloud correspondences, our method uses a robust metric to quantify the goodness of alignment of two point clouds. We shall explain this shortly. For clarity we start with a description of the process using the simpler Euclidean distance based metric which captures the spirit of the process. In this case, the three terms in \eqref{e:objFn} can be written as follows:
\begin{align}
  E_{match} =& E_\text{match-fast} + E_\text{match-accurate}  \label{e:match}\\
            =& \sum_{i=1}^{n} (\mathbf{n}_i^T (\mathbf{z}_i-C_Y(\mathbf{z}_i) ))^2+  \label{e:pt2pl}\\
             & \sum_{i=1}^{n} (\mathbf{z}_i-C_Y(\mathbf{z}_i))^2  \label{e:pt2pt}\\
  E_{rigid} =& \sum_{i=1}^{n} {\|\mathbf{z}_i-(\mathbf{R}\mathbf{x}_i+\mathbf{t})\|_2}^2  \label{e:rigid}\\
  E_{model} =& \sum_{i=1}^{n} {\|\mathbf{z}_i-(\mathbf{P}_i\mathbf{d}+\mathbf{m}_i)\|_2}^2. \label{e:model}
\end{align}
Here $Z=\{z_1,\ldots,z_n\}$ is a deformed point cloud $X$ (hence for all indexes $i$, the point $z_i$ corresponds to $x_i$) which is being aligned with $Y$, $\mathbf{R}$ and $\mathbf{t}$ respectively the rotation matrix and the translation vector which describe the rigid transformation of the source point cloud, $\mathbf{P}_i$ and $\mathbf{m}_i$ the parts of respectively the matrix of morphable model principal components of shape and the mean shape, $C_Y(\mathbf{z}_i)$ the point in  the target point cloud closest to $\mathbf{z}_i$, and $\mathbf{n}_i$ the surface normal at $C(\mathbf{z}_i)$. In words, the term $E_\text{match}$ comprises two contributions: $E_\text{match-fast}$ and $E_\text{match-accurate}$. The first of these can be seen to accumulate point-to-plane errors between the point cloud $Z$ and the surface described by $Y$. For reasons of efficiency this has in the past been used as a linearized version of the point-to-point error of $E_\text{match-accurate}$. However, we found that the inclusion of both $E_\text{match-fast}$ and $E_\text{match-accurate}$ provided the best trade-off between the two. Continuing with the term in \eqref{e:rigid}, $E_\text{rigid}$ penalizes large non-rigid deformations between $X$ and $Z$ i.e.\ differences between different $x_i$ and $z_i$ which cannot be explained by simple global rotation and translation. Lastly $E_{model}$ can be understood as implementing the distance-from-feature-space metric~\cite{AranCipo2006e,Aran2014,WangShanChenDai+2012} where the feature space is spanned by the morphable model principal components of shape; the further the shape described by $Z$ is from the best reconstruction by the morphable model, the greater the corresponding penalty is.

The optimization problem described by \eqref{e:objFn} is solved iteratively. In particular the first step is to linearize the updates to the rotation matrix using first order Taylor expansion -- since the updates are by their very nature assumed small (this is particularly true in our algorithm given that the localization of facial features described in the previous sections allows us to initialize the model well) all cosine terms are approximated by 1 and all sine terms by the corresponding angles. This results in the rotation update matrix $\widetilde{\mathbf{R}}$ of the form:
\begin{align}
  \widetilde{\mathbf{R}} = \begin{bmatrix}
                             ~~~1 & ~~~a & ~~~b \\
                               -a & ~~~1 & ~~~c \\
                               -b &   -c & ~~~1 \\
                           \end{bmatrix}.
\end{align}
Thus the iterative process can be described by the following equation:
\begin{align}
  \arg \min_{Z^{k+1}, \textbf{d}, \widetilde{R}, \widetilde{t}} \sum_{i=1}^n \Bigg\{
  &(\mathbf{n}_i^T(\mathbf{z}_i^{k+1}- C_Y(\mathbf{z}_i)^k))^2 + \notag \\
  &(\mathbf{z}_i^{k+1}-C_Y(\mathbf{z}_i^k))^2 + \notag \\[7pt]
  &\omega_1~{\|\mathbf{z}_i^{k+1}-(\widetilde{\mathbf{R}}(\mathbf{R}\textbf{x}_i+\mathbf{t})+\widetilde{\mathbf{t}})\|_2}^2+ \notag \\[7pt]
  &\omega_2~{\|\mathbf{z}_i^{k+1}-(\mathbf{P}_i\textbf{d}+\mathbf{m}_i)\|_2}^2 \Bigg\}
  \label{e:objFnLin}
\end{align}
where $k$ is the iteration number which modifies each of the iteratively updated variables to denote their values in the corresponding iteration (such as $Z^{k+1}$ for example), and $\widetilde{\mathbf{R}}$ and $\widetilde{\mathbf{t}}$ are respectively the update to the rotation matrix and the translatory adjustment between the source and the target. The iteration is initialized with $Z^0 = X$ and, as we explained before, $\mathbf{R}$ and $\mathbf{t}$ computed from the 3D loci of the detected facial features. Notice that in the computation of the closest points in $Y$ to each $\mathbf{z}_i$, for tractability reasons it is the previous set of estimates $Z^k$ that is being used rather than $Z^{k+1}$ i.e.\ $C_Y(\mathbf{z}_i^k)$ instead of $C_Y(\mathbf{z}_i^{k+1})$.

\subsubsection{Increasing fitting robustness}\label{sss:robust}
When applied on real-world data, the model fitting error function \eqref{e:objFn} formulated using the Euclidean distance metrics \eqref{e:pt2pl}--\eqref{e:model} is readily found to exhibit difficulties posed by noise and incorrect matches between two point clouds. As already noted, the sensed depth data is highly noisy so this is a major practical challenge. On the other hand the latter problem of incorrect matches may occur when there are missing point cloud data such as when a part of the sensed surface is occluded. In this case in \eqref{e:pt2pl} a model point (recall: obtained by sampling from the surface generated by the 3D morphable model) may be matched with a point which corresponds to an entirely different (and hence incorrect) part of the face surface. The Euclidean metric allows such misleading matches to contribute greatly to the overall error function thereby misleading the iterative process. This has negative consequences both to the accuracy of the fit as well as the efficiency of the fitting process i.e.\ its speed of convergence.

To dampen the effect of noisy and incorrect matches we employ a robust metric $\psi$ to modulate the contribution of each term in the summations in \eqref{e:pt2pl} and \eqref{e:pt2pt}:
\begin{align}
E&_{match}' = E_\text{match-fast}' + E_\text{match-accurate}'  \label{e:matchRob}\\
            =& \sum_{i=1}^{n} \psi(| \mathbf{n}_i^T (\mathbf{z}_i-C_Y(\mathbf{z}_i) ) |)~(\mathbf{n}_i^T (\mathbf{z}_i-C_Y(\mathbf{z}_i) ))^2+  \label{e:pt2plRob}\\
             & \sum_{i=1}^{n} \psi(\| \mathbf{z}_i-C_Y(\mathbf{z}_i) \|_2)~(\mathbf{z}_i-C_Y(\mathbf{z}_i))^2  \label{e:pt2ptRob}.
\end{align}
For $\psi$ we use Tukey's biweight function~\citep{HubeRonc2009}:
\begin{align}
  \psi(d) = \begin{cases}
               1-\left( d/d_t \right)^2 & d \leq d_t\\
               0       & d > d_t
  \end{cases},
\end{align}
where $d$ is an input distance argument such as $| \mathbf{n}_i^T (\mathbf{z}_i-C_Y(\mathbf{z}_i) ) |$ in \eqref{e:pt2plRob} or $\| \mathbf{z}_i-C_Y(\mathbf{z}_i) \|_2$ in \eqref{e:pt2ptRob}, and $d_t$ the threshold which governs the breadth of the function's influence. We used $d_t=0.01$ which corresponds to the physical distance of $0.01$~m.  The iterative process summarized by the expression in \eqref{e:objFnLin} remains unchanged so we do not repeat the equation which is understood to include the described weighting terms.

\subsubsection{Adaptive descent}\label{sss:reweighting}
The three terms in \eqref{e:objFn} differ substantially in terms of their ability to compensate for fitting errors of different magnitudes. While the point cloud matching term \eqref{e:match} and the rigidity-constraining term \eqref{e:rigid} can effect fitting parameter changes which cross large spatial distances, the term \eqref{e:model} which corresponds to the goodness of fit of the 3D morphable model is far more spatially constrained. This is a consequence of relatively small variability of faces and in particular their shape~\citep{CrawCostKatoAkam1999,GrosYangWaib2000}. Hence even the faces of different individuals are sufficiently similar to be registered well using a rigid transformation only.

The aforementioned observations lend a useful insight. Firstly, from the point of computational efficiency, if all of the terms in \eqref{e:objFn} are included from the very start of the fitting process, resources are unnecessarily wasted on the estimation and updating of the 3D morphable model parameters -- if the rigid registration parameters are too far from their optimum values, the tuning of the model which describes intricate inter-personal differences is not done with sufficiently good data. Secondly, the misguided adjustments of the 3D morphable model parameters which occur in the early stages of the fitting process can accumulate and make the final fitting stages (when only fine refinements of the rigid registration may be needed) excessively slow and possibly produce a result of lower accuracy than one which would be achieved if no prior adjustments had been made.

Guided by the analysis above we implement an error term re-weighting scheme which at the same time achieves fast and robust convergence, and accurate fitting. In particular we start the fitting process with the model error term \eqref{e:model} entirely suppressed by setting $\omega_2$ to 0 -- we shall denote this initial weight as $\omega_2^{(0)}=0$. When the error function \eqref{e:objFn} reaches a local minimum we declare that the first stage of the process is complete and that the generic shape model is approximately registered with the target. At this point the value of $\omega_1$ is gradually reduced with a concurrent increase of $\omega_2$ from its initial value of $\omega_2^{(0)}=0$. The changes to $\omega_1$ and $\omega_2$ stop when $\omega_1$ reaches a preset minimum value $\omega_1^\infty$ and when $\omega_2$ reaches its preset maximum $\omega_2^\infty$. The fitting process itself continues until convergence.

\subsection{Gaze from synthetically generated images}
After the fitting of a morphable 3D shape model to the sensed depth data, both the inherent (i.e.\ person specific) and relative (i.e.\ pose specific) geometric configurations are known and accessible explicitly. Moreover the complementary RGB information can be readily used to associate a texture map with the inferred face shape. We use these observations to render a synthetic frontal face image by normalizing the head pose i.e.\ by simulating a rigid transformation of the head which places it in front of and facing the camera. This is the second step in the pipeline shown in Figure~\ref{f:overview}.

\begin{figure}[htb]
  \centering
  \includegraphics[trim={3cm 0 2cm 0},clip,width=\columnwidth]{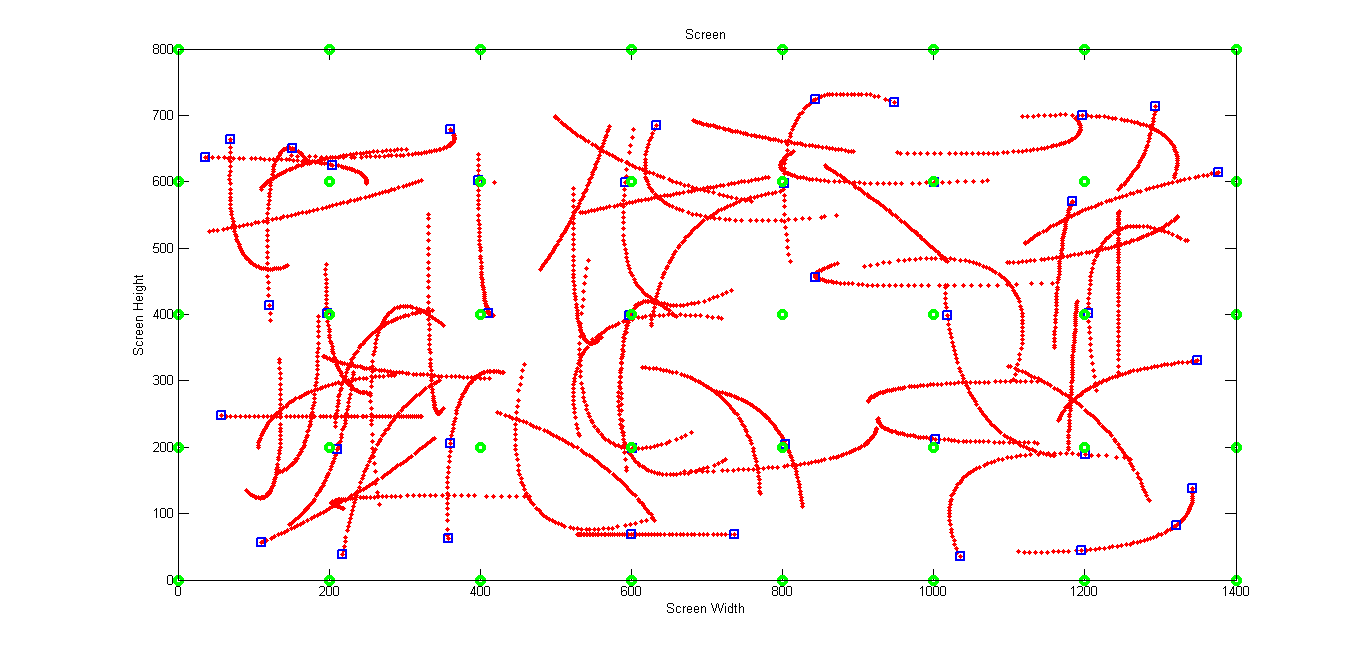}
  \caption{ A sparse set of training points of gaze and the corresponding eye appearance images are chosen from the original gazing tracks (red) by sampling the screen space uniformly (green points), and using only those points (blue) from the original gaze tracks closest (in the Euclidean distance sense) to the sparse samples.}
  \label{f:ALRtrain}
\end{figure}

Hereafter the steps performed by our algorithm follow under the umbrella of appearance based estimation methods discussed in Section~\ref{s:intro}. However at this stage it is important to highlight a few important distinctions. Firstly having obtained a highly accurate 3D model of the face our algorithm does not need to inter the locations of the eyes. Rather these are known \textit{a priori} seeing that the 3D morphable model comprises mixing shapes which are semantically mutually co-registered and annotated. This allows us to extract eye appearance images with high accuracy and effectively perfect reliability.

The second important effect of the preceding pose normalization step is that the learning space is greatly reduced. In particular, while the existing `conventional' appearance based gaze estimation algorithms have to learn the mapping from eye appearance to gaze direction space over a broad range of different head poses and relative pupil loci, having normalized pose using photorealistic 3D rendering our learning is constrained to learning pupil movement only. This makes the learning process both inherently easier and in terms of practical demands reduces the amount of data required to perform the aforesaid learning.

\subsection{Feature extraction}
Thus this second stage of our algorithm begins by extracting greyscale eye appearance patches. As mentioned earlier this is readily achieved because the locations of the eyes are explicitly given by our 3D morphable model. To reduce the dimensionality of the representation we downsample the patches to the uniform scale of $3 \times 5$ pixels thus obtaining 15D feature vectors. Previous work suggests that this scale is sufficient for accurate gaze estimation; our results presented in the next section corroborate this finding.

\subsection{Learning the appearance to gaze mapping}
In principle any of a number of regression approaches can be applied at this stage. For the sake of comparison we chose to adopt two well known, widely used, and well understood methods of different complexities. These are (i) simple $k$-nearest neighbour (kNN) regression, and (ii) adaptive linear regression (ALR) \citep{LuSugaOkabSato2011}. These are summarized briefly next.

\paragraph{$k$-nearest neighbour regression} As other $k$-nearest neighbour based algorithms \citep{KhanAhma2004,Aran2013c}, kNN regression is a non-parametric technique. Given an input independent variable $x$ (in our case this is an eye appearance image), the corresponding dependent variable value $y$ (in our case this is gaze direction) is predicted by finding the $k$-nearest neighbours $x_{i(1)},\ldots,x_{i(k)}$ to $x$ in the training set, and then by computing a weighted summation of the dependent variable values $y_{i(1)},\ldots,y_{i(k)}$ associated with them:
\begin{align}
  y = \sum_{j=1}^k w_j y_{i(j)},
\end{align}
where the values of the weight $w_j$ is inversely proportional to the distance between $x$ and $x_{i(j)}$:
\begin{align}
  w_j = \| x - x_{i(j)} \|_2^{-1}.
  \label{e:knnweight}
\end{align}
As per \eqref{e:knnweight} we used the Euclidean distance though any of a number of alternatives, such as the Minkowski distance, could be employed just as readily.

\paragraph{Adaptive linear regression (ALR)}
Linear regression relates an input independent variable $x$ with the corresponding output dependent variable $y$ through a linear transformation:
\begin{align}
  y=Ax.
  \label{e:alr}
\end{align}
Adaptive linear regression draws from this idea and the observation that if the number of training samples is greater than the dimensionality of the independent variable, a more input specific mapping can be found by exploiting the structure of the input space. Specifically in the present case images of eyes can be considered to lie approximately on what somewhat loosely may be described as an eye manifold. Much like the better known face manifold \citep{LeeHoYangKrie2003a,LuiBeve2008,WangShanChenDai+2012}, the eye manifold is approximately smooth and highly non-linear. Therefore, rather than learning the global projection matrix $A$ in \eqref{e:alr}, adaptive linear regression adaptively learns this mapping for the specific sample of interest i.e.\ for the specific region of the independent variable space. Relevant training samples from which learning is performed are chosen on the basis of the criterion which attempts to maximize the linear representability of the input sample. Full detail on this technique can be found in the work of \cite{LuSugaOkabSato2011}. Following their work we used a sparse training set, as illustrated in Figure~\ref{f:ALRtrain}.

\section{Evaluation}
For the evaluation of the proposed method and its comparison with the state of the art we adopted the well known EYEDIAP database~\citep{FuneOdob2012} \footnote{The database can be downloaded from \url{http://www.idiap.ch/dataset/eyediap}.}. It is a freely and publicly available standard benchmark for the evaluation of algorithms for gaze estimation from RGB-D data. A detailed description of the database, and the protocol used for its acquisition and ground truth labelling can be found in the original paper of \cite{FuneOdob2012}; for completeness herein we summarize the key aspects of the database of relevance to the present work.

EYEDIAP contains RGB-D sequences acquired using Microsoft Kinect at VGA resolution of $640 \times 480$ pixels and at 30~fps. The total number of individuals in the database is 16, of which 12 are male and 4 female. Each user participated in multiple acquisition sessions of 2 to 3 minutes, resulting in the total number of 94 sequences (total duration of over 4 hours) across the database. Of particular interest to the problem addressed in the present work is that the control over head motion which was imposed during data acquisition. Specifically, in some sessions the users were asked to track a screen target while keeping the head still, while in others natural head movement was not constrained. Because the location of the screen target was controlled by the experimenters and its tracking was not challenging (its motion was not excessively fast or erratic) the ground truth is considered \textit{ipso facto} known.

\subsection{Results and discussion}
We start our analysis of the experimental results by comparing the performance of the proposed method with that of \cite{FuneOdob2012} on the less challenging subset of videos in the EYEDIAP database, in which as we noted previously the users kept their head stationary and effected gaze changes by means of eye movement only. A summary of the key findings can be found in Tables~\ref{t:resKNNstat} and~\ref{t:resALRstat}. Considering that some of the key strengths of the proposed method are effected by the highly accurate 3D face model fitting, we found it rather surprising that even though in this simpler challenge the required pose normalization was small in extent, our method already exhibited significantly superior performance than the state of the art method of \cite{FuneOdob2012}. When simple kNN regression was used the average reduction in the error of the gaze direction estimate was approximately 22\% (being 7.7$^\circ$ in comparison with 9.9$^\circ$, see Table~\ref{t:resKNNstat}); even better performance (average error of 7.2$^\circ$, see Table~\ref{t:resALRstat}) and even greater reduction (nearly 30\%) in the estimate error was attained with the application of the more complex adaptive linear regression.

\begingroup\setlength{\fboxsep}{0pt}
\begin{table}
  \centering
  \caption{Gaze direction estimate errors obtained using $k$-nearest neighbour regression on the EYEDIAP subset of video sequences in which the users were instructed to keep their head still and alter their gaze by means of eye movement only.}\vspace{8pt}
  \renewcommand{\arraystretch}{1.5}
  \colorbox{lightgray}{%
  \begin{tabular}{l!{\color{white}\vrule}cc!{\color{white}\vrule}c}
  \Hline
                    & Left eye    & Right eye    & Mean\\
    \arrayrulecolor{white}\hline
    Proposed method & 8.8$^\circ$ & 6.5$^\circ$  & 7.7$^\circ$\\
    Previous state of the art           & 10.2$^\circ$ & 9.6$^\circ$ & 9.9$^\circ$\\
    \Hline
  \end{tabular}}
  \label{t:resKNNstat}
\end{table}

\begingroup\setlength{\fboxsep}{0pt}
\begin{table}
  \centering
  \caption{Gaze direction estimate errors obtained using adaptive linear regression on the EYEDIAP subset of video sequences in which the users were instructed to keep their head still and alter their gaze by means of eye movement only. }\vspace{8pt}
  \renewcommand{\arraystretch}{1.5}
  \colorbox{lightgray}{%
  \begin{tabular}{l!{\color{white}\vrule}cc!{\color{white}\vrule}c}
  \Hline
                    & Left eye    & Right eye    & Mean\\
    \arrayrulecolor{white}\hline
    Proposed method & 7.5$^\circ$ & 6.9$^\circ$  & 7.2$^\circ$\\
    Previous state of the art           & 9.7$^\circ$ & 10.5$^\circ$ & 10.1$^\circ$\\
    \Hline
  \end{tabular}}
  \label{t:resALRstat}
\end{table}

Following the highly promising results obtained already on the simpler task of gaze estimation from video sequences in which the users' were asked to keep their heads still, we next compared our method with the state of the art on the more challenging and more practically relevant problem of estimating and tracking gaze direction when natural head movement accompanied the movement of the eyes. As previously, we summarize the key results in Tables~\ref{t:resKNNmov} and~\ref{t:resALRmov}. In line with our expectations both the proposed method and that of Funes and Odobez performed less well in this less constrained setup. This is witnessed by the increase in the gaze direction error. However, importantly, it can be readily observed that the aforesaid error increase is rather different across the two methods. For example, looking at the results obtained using kNN regression, it can be seen that the error of the proposed method increased from the previous value of 7.7$^\circ$ to 8.9$^\circ$ i.e.\ for approximately 15\%. In contrast, the error of the estimates achieved by the algorithm of Funes and Odobez increased from 9.9$^\circ$ to 16.3$^\circ$ which corresponds to a far greater proportional error increase of approximately 65\%. Comparing the average errors of the methods directly shows that the average error of the proposed method is nearly half that of the previous state of the art (8.9$^\circ$ compared with 16.3$^\circ$) using simple kNN regression. Somewhat smaller but still major improvement of 36\% is observed with the use of the adaptive linear regression.

\begingroup\setlength{\fboxsep}{0pt}
\begin{table}
  \centering
  \caption{Gaze direction estimate errors obtained using $k$-nearest neighbour regression on the EYEDIAP subset of video sequences in which the users were allowed to move their head naturally while following a target displayed on the screen. }\vspace{8pt}
  \renewcommand{\arraystretch}{1.5}
  \colorbox{lightgray}{%
  \begin{tabular}{l!{\color{white}\vrule}cc!{\color{white}\vrule}c}
  \Hline
                    & Left eye    & Right eye    & Mean\\
    \arrayrulecolor{white}\hline
    Proposed method & 9.0$^\circ$ & 8.9$^\circ$  & 8.9$^\circ$\\
    Previous state of the art & 18.0$^\circ$ & 14.6$^\circ$ & 16.3$^\circ$\\
    \Hline
  \end{tabular}}
  \label{t:resKNNmov}
\end{table}

\begingroup\setlength{\fboxsep}{0pt}
\begin{table}
  \centering
  \caption{Gaze direction estimate errors obtained using adaptive linear regression on the EYEDIAP subset of video sequences in which the users were allowed to move their head naturally while following a target displayed on the screen. }\vspace{8pt}
  \renewcommand{\arraystretch}{1.5}
  \colorbox{lightgray}{%
  \begin{tabular}{l!{\color{white}\vrule}cc!{\color{white}\vrule}c}
  \Hline
                    & Left eye    & Right eye    & Mean\\
    \arrayrulecolor{white}\hline
    Proposed method & 9.8$^\circ$ & 9.5$^\circ$  & 9.6$^\circ$\\
    Previous state of the art           & 15.6$^\circ$ & 14.2$^\circ$ & 14.9$^\circ$\\
    \Hline
  \end{tabular}}
  \label{t:resALRmov}
\end{table}

\section{Summary and conclusions}\label{s:summary}
In this paper we described a novel algorithm for gaze direction estimation from RGB-D data acquired using an affordable, consumer market device such as Microsoft Kinect. The method we introduced comprises two key stages. In the first stage accurate head pose estimation is achieved by fitting a person specific morphable model of the face to depth data. Our approach achieves high accuracy and high speed through a carefully engineered fitting objective function which comprises a combination of terms which excel either in accurate or fast point cloud matching. The contribution of these terms is then adaptively adjusted during the iterative process of model fitting i.e.\ model parameter estimation. Following the fitting of an accurate 3D face model, pose normalization is done by re-rendering the model from the frontal view. In the second stage of the proposed method, appearance based eye features are extracted from the synthetic image and used to train a regressor. The proposed algorithm was evaluated on the standard benchmark database EYEDIAP on which it is shown to outperform significantly the current state of the art, reducing the error in the gaze direction estimate by more than a third.

\section{Acknowledgements}
The authors would like to thank Prof.\ Denis Laurendeau for sharing his thoughts and experienced through numerous discussions related to the work described in the present paper.
%\clearpage

\balance

%\small
\bibliographystyle{named}
\bibliography{../../../my_bibliography}

\end{document}